\documentclass[letterpaper, 10 pt, conference]{ieeeconf}  % Comment this line out if you need a4paper

\usepackage{amsmath,amssymb,amsfonts}
\usepackage{graphicx}
\usepackage[colorlinks,bookmarksopen,citecolor=green,urlcolor=red]{hyperref}
\usepackage{subcaption}
\usepackage{booktabs}
\usepackage{makecell}

\usepackage{xcolor}
\usepackage[table, svgnames, dvipsnames]{xcolor}

\def\eg{\emph{e.g.}}

\def\etc{\emph{etc.}} 
\def\etal{\emph{et al.}}

\IEEEoverridecommandlockouts                              % This command is only needed if                                         % you want to use the \thanks command

\title{\LARGE \bf
SubSense: VR-Haptic and Motor Feedback for Immersive Control\\ in Subsea Telerobotics
\vspace{-4mm}
}

\author{Ruo Chen, David Blow, Adnan Abdullah, Md. Jahidul Islam$^\diamond$% <-this % stops a space
\\
{RoboPI Laboratory, Department of ECE, University of Florida, FL, US}
\thanks{\noindent\rule{8cm}{0.6pt}}
\thanks{$^\diamond$Presented at the OCEANS 2025 Great Lakes Conference.}
\thanks{Webpage: \url{https://robopi.ece.ufl.edu/subsea_tr.html}}
\vspace{-1mm}
}

\begin{document}

\maketitle
\thispagestyle{empty}
\pagestyle{empty}

%\input{src/oceans}
%\newpage

%%%%%%%%%%%%%%%%%%%%%%%%%%%%%%%%%%%%%%%%%%%%%%%%%%%%%%%%%%%%%%%%%%%%%%%%%%%%%%%%
\begin{abstract}
% Version 2
This paper investigates the integration of haptic feedback and virtual reality (VR) control interfaces to enhance teleoperation and telemanipulation of underwater ROVs (remotely operated vehicles). Traditional ROV teleoperation relies on low-resolution 2D camera feeds and lacks immersive and sensory feedback, which diminishes situational awareness in complex subsea environments. We propose SubSense -- a novel VR-Haptic framework incorporating a non-invasive feedback interface to an otherwise 1-DOF (degree of freedom) manipulator, which is paired with the teleoperator's glove to provide haptic feedback and grasp status. Additionally, our framework integrates end-to-end software for managing control inputs and displaying immersive camera views through a VR platform. We validate the system through comprehensive experiments and user studies, demonstrating its effectiveness over conventional teleoperation interfaces, particularly for delicate manipulation tasks. Our results highlight the potential of multisensory feedback in immersive virtual environments to significantly improve remote situational awareness and mission performance, offering more intuitive and accessible ROV operations in the field.
\end{abstract}

%%%%%%%%%%%%%%%%%%%%%%%%%%%%%%%%%%%%%%%%%%%%%%%%%%%%%%%%%%%%%%%%%%%%%%%%%%%%%%%%
\section{Introduction}
Remotely Operated Vehicles (ROVs) are indispensable tools in the marine industry, offering a safer and more cost-effective alternative to human divers~\cite{abdullah2024human}. Underwater ROVs are versatile platforms supporting a range of missions, from routine imaging and infrastructure inspection to complex tasks such as environmental monitoring~\cite{aguirre2019design}, maintaining subsea infrastructure~\cite{jacobi2015autonomous,patel2024multi}, performing mine countermeasure and explosive ordinance disposal~\cite{reed2010detection}, salvaging, search-and-rescue~\cite{matos2016multiple}, and deep-water expeditions~\cite{howard2006experiments}. With over $79\%$ of subsea deployments done by ROVs, they play a crucial role in commerce, military, and science -- enabling us to explore beyond the limits of human scuba divers~\cite{teague2018potential}. 

Despite growing industrial demands and recent advancements, underwater ROVs still have inherent limitations, particularly in their immersive control and interaction capabilities. Commercial teleoperation interfaces predominantly rely on 2D visual feeds and joystick controllers, restricting feedback to purely visual cues and relying heavily on their trained skills and intuitions for safe operation~\cite{jamieson2013deep}. This stands in stark contrast to divers, who benefit from direct multisensory feedback in underwater environments~\cite{riley2016situation}. The absence of tactile and situational sensory information poses significant challenges, particularly in low-light and confined spaces such as underwater caves~\cite{abdullah2023caveseg}, where operators are at heightened risk of spatial disorientation and collisions~\cite{abdullah2024ego2exo}.

Similar issues arise in open-water scenarios, where ocean currents can induce disorientation and drift~\cite{hapticsubseanavXia2021}. The lack of haptic feedback further complicates tasks requiring precise manipulation, such as maintenance and repair, increasing the risk of operational errors~\cite{kampmann2015towards}. This fundamental disconnect between the teleoperator and the ROV's surroundings is well-documented in recent studies~\cite{abdullah2024ego2exo,xia2023rov}, as well as corroborated by experiences in our own field trials; highlighting the challenges of deploying in deep-water environments and performing manipulation tasks~\cite{rahal2020caring}. 

\begin{figure}[t]
    \centering
    \includegraphics[width=0.98\linewidth]{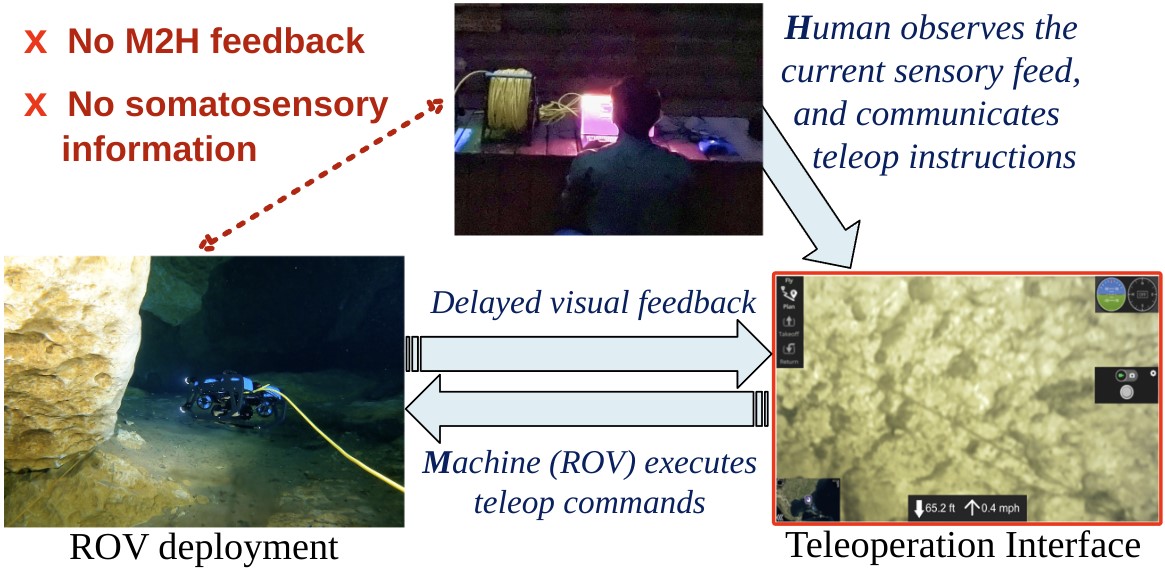}%
    \vspace{-1mm}
    \caption{Snapshots from a subsea inspection scenario are shown, where the human has to make real-time teleoperation decisions based on noisy visual feedback alone. From our \underline{field trials in underwater caves}, we found that limited \textbf{M2H} (Machine to Human) feedback with no somatosensory information is a major limitation of existing teleoperation interfaces, which we attempt to address in SubSense.}%
    \label{fig:challenge}
    \vspace{-4mm}
\end{figure}

For any \textit{motor action}, humans rely on somatosensory and visual feedback to interpret the results of actions within their environment~\cite{ImpactofVisualWood2013,PerceptualMotorJax2009}. Compared to ROVs, a diver in the underwater environment can more easily perform delicate manipulation tasks due to being fully engaged with their senses. As shown in Figure~\ref{fig:challenge}, when limited to only visual feedback, the difficulty increases due to the lack of supplementary information such as tactile or haptic feedback~\cite{MethodsHapticFeedbackOkamura2004}. Thus, ROV teleoperators lack the sensory information needed to grasp objects when more precision is required~\cite{GripForceTeleroboticsKhurshid2017, ExploratoryMovementsDebats2010}.

\begin{figure*}[t]
    \centering
    \includegraphics[width=0.48\linewidth]{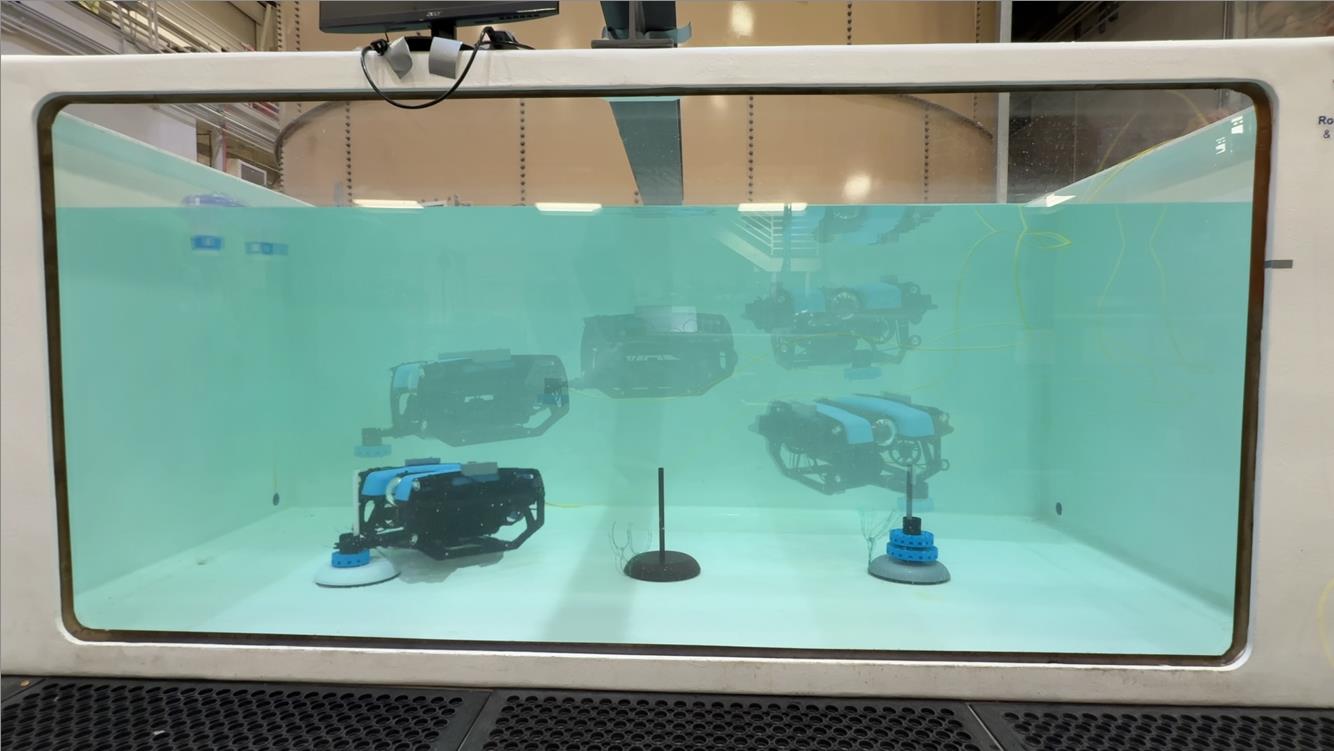}
    \includegraphics[width=0.32\linewidth]{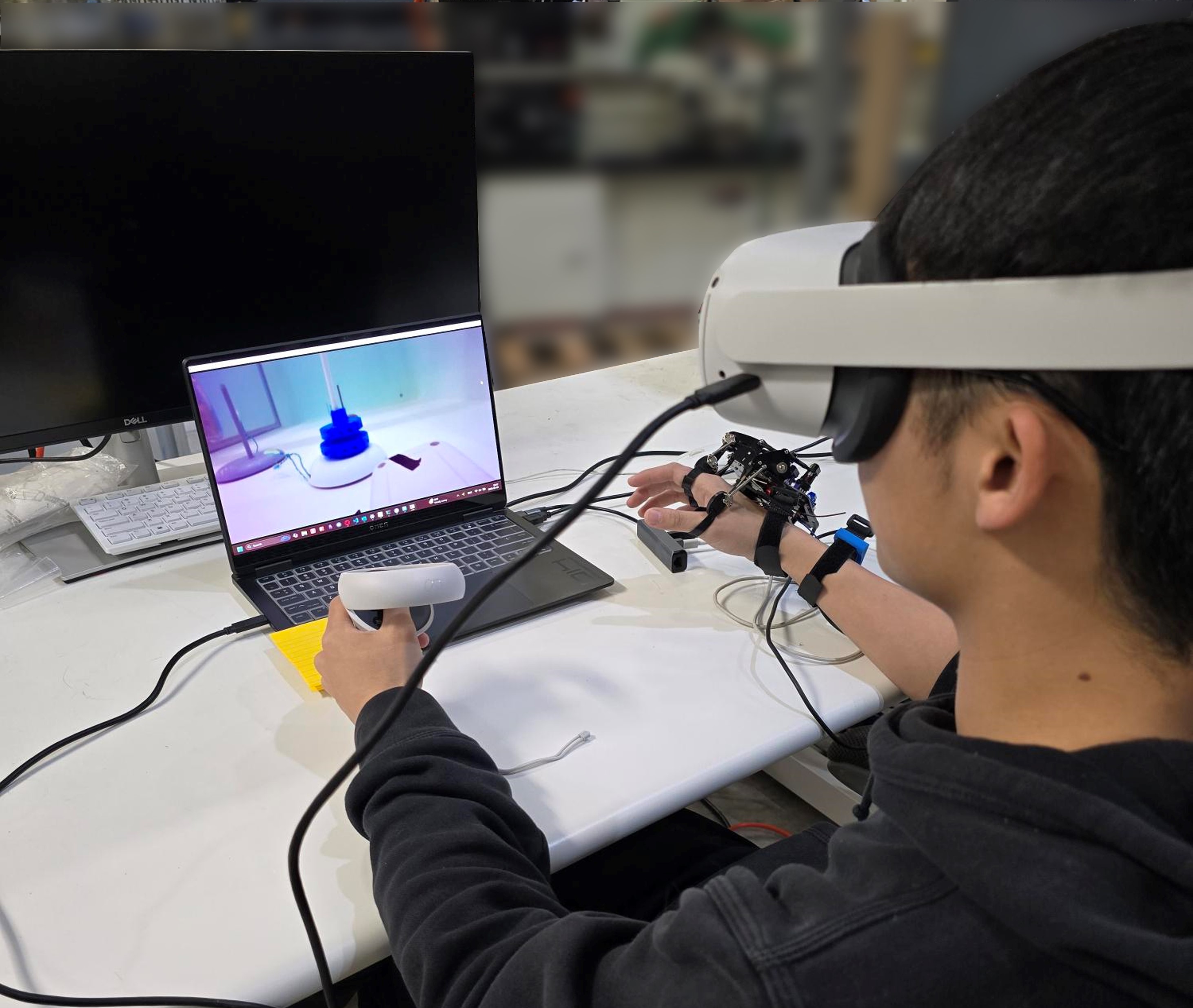}~
    \caption{A 3-disk Tower of Hanoi (ToH) setup where participants are asked to move the ordered disk from one pole to another using an intermediate pole. (Left) first subtask of the ROV: moving the topmost disc; (right) Teleoperator is using SubSense to achieve it. A sample demo of the task being completed can be viewed here: \url{https://youtu.be/0tOXA9pts_Y}.}
    \label{fig:banner}
    \vspace{-4mm}
\end{figure*}

Recent advancements in virtual reality (VR) technologies have opened new avenues for improving human-machine interactions in remote environments~\cite{Hirschmanner_VRTeleoperation,hine1994application}. This indicates the potential to significantly enhance ROV teleoperation by enabling immersive experiences for teleoperators through a simulated \textit{digital twin}~\cite{abdullah2024human,McHenry2021_PredictiveXR}. When combined with haptic feedback, VR can simulate a more embodied experience to better understand spatial relations within their environment~\cite{xia2024human,vasilijevic2012acoustically}, assisting with tasks that require precise manipulation, \eg, handling sensitive samples.

While the integration of sensory feedback has been extensively studied in fields such as industrial and surgical robotics~\cite{SurgicalHapticsPatel2022,IndTeleropationGonzalez2021}, the application of haptic feedback in subsea telemanipulation tasks remains largely underexplored~\cite{abdullah2024human}. Furthermore, only a limited number of user studies have examined the impact of subsea ROV teleoperation through VR in physical environments~\cite{xia2022virtual,elor2021catching}. Among these, most evaluations are conducted in simulation, lacking validation in real-world scenarios~\cite{ryden2013advanced,xia2023visual}. This underscores a significant gap, which we demonstrated in~\cite{abdullah2024ego2exo,gupta2025demonstrating} through field experimental trials for ROV-based mapping and inspection tasks inside underwater caves. 

%in the integration of haptic feedback and VR-based teleoperation within marine robotics. In, we demonstrated these issues in details through 

%%%%%%%%%%%%%%%%%%%%%%%%%%%%%%%%%%%%%%%%%%%%%%%%%%%%%%%%%%%%%%%%%%%%%%%%%%%%%%%%%%%%%%%%%%%%%%%%%%%%%%%%%%%%%%%%%%%%%%%%%%%%%%%%%%%%%%%%%%%%%%%%%%%%

% \vspace{1mm}
% \noindent
% \textbf{Contributions}
In this paper, we present \textbf{SubSense}, a VR-haptic framework designed to enhance teleoperation by immersing the operator in the ROV environment. This is achieved through the integration of haptic feedback, body movement-based ROV control, and a head-mounted display (HMD) providing a real-time video feed from the ROV’s onboard camera. The system utilizes a BlueROV2 equipped with a Newton Subsea gripper, which inherently supports only open-loop commands (neutral, open, close). To enable closed-loop control, we integrate a \textbf{non-invasive augmentation for gripper position} feedback, allowing the manipulator’s position to be mapped to a \textbf{user-worn glove} while providing vibration feedback upon successful object grasping. Additionally, we integrate the system with a commercial VR platform, enabling seamless ROV control and real-time video streaming within the HMD to enable human-machine-embodied teleoperation.

We assess human embodiment through a user study involving eight participants, evaluating their performance in transferring delicate objects between predefined locations within a controlled physical testing environment. Specifically, we use a 3-disk Tower of Hanoi (TOH) experimental setup, as shown in Figure~\ref{fig:banner}.  Participants operate the ROV in two modes: with (\textbf{1}) traditional first-person view (FPV); and (\textbf{2}) VR-based control with haptic feedback through our SubSense framework. Performance is quantified based on task completion time and the number and severity of object damages. Additionally, participants complete a post-trial survey to subjectively evaluate satisfaction and ease of use for each mode of operation. Our findings indicate that immersive interfaces enhance remote situational awareness and operator embodiment in underwater ROV teleoperation. 

While preliminary, these results provide a strong foundation for further development and larger-scale evaluations in more complex environments. Ongoing work focuses on improving the user experience through (\textbf{i}) reducing latency by optimizing the communication and rendering performance to maintain a responsive experience; (\textbf{ii}) upgrades to the haptic system to deliver tactile cues that better reflect the physical interaction; (\textbf{iii}) refinements to the control scheme to allow for more intuitive ROV movement; and (\textbf{iv}) incorporation of additional data streams from to the user such as ROV orientation, speed, and other interactive information.

\begin{figure*}[t]
    \centering    \includegraphics[width=0.96\linewidth]{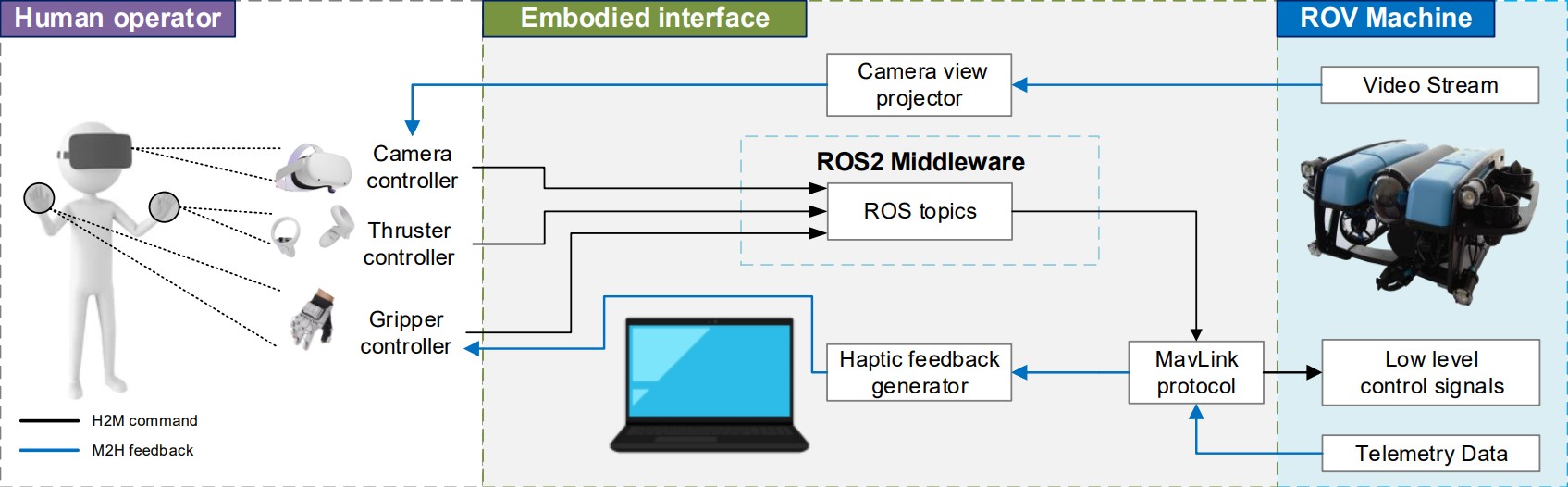}%
    \vspace{-1mm}
    \caption{Overview of the proposed multi-modal ROV teleoperation system is shown. The operator embodies the ROV via haptic feedback and a rendered camera feed to control its motion, camera orientation, and gripper pose. The sensorimotor  via a controller, head movement, and hand glove.
    }
    \label{fig:system}
    \vspace{-4mm}
\end{figure*}

\section{Background and Related Work}
\subsection{Subsea Telerobotics and Tele-manipulation}
Subsea telerobotic technologies are widely used for remote inspection, intervention, and maintenance of natural and man-made subsea infrastructures including oil and gas rigs, pipelines, umbilicals, coral reefs, \etc~\cite{maurelli2016pandora,petillot2019underwater,patel2024multi}. These operations involve manipulating underwater objects with the ROV's arm/gripper, which is typically controlled by a human operator through direct joint-by-joint teleoperation method~\cite{jamieson2013deep}. However, the precision of such systems is constrained by limited visibility~\cite{islam2024eob}, low communication bandwidth~\cite{yang2022position}, inaccurate localization~\cite{kim2020cooperative}, lack of intuitive feedback~\cite{khedr2024design}, and high cognitive demand~\cite{zhang2024adaptive,teigland2020operator}.

Researchers have developed several feedback mechanisms and interactive control systems to assist teleoperators in subsea tele-manipulation~\cite{abdullah2024human}. Traditional hand-held controllers (\eg, joystick) are replaced by more natural commands, including hand gestures and body motions~\cite{gancet2015dexrov,phung2024shared}. These gestures, captured by cameras~\cite{kapicioglu2021touchless}, hand-mounted IMUs~\cite{babiarz2024underwater}, finger-mounted magnetic sensors, and body suits~\cite{xia2023rov}, are mapped to specific commands for manipulation. In addition to improved control methods, multi-modal feedback techniques engage the operator's sensory perception by mimicking the ROV's interaction with its environment~\cite{xia2023sensory}. Moreover, interactive vision systems such as a camera paired with the operator's head movement~\cite{sobhani2020robot}, augmented viewpoints~\cite{abdullah2024ego2exo}, and depth perception~\cite{bruno2018augmented} increase grasping accuracy in complex scenarios.

\subsection{XR and Haptics Integration for ROV Teleoperation}
The extreme underwater environments and the limitations of maneuvering ROVs using video streams alone have driven the development of augmented visuals to provide more peripheral information~\cite{bruno2018augmented,xu2021vr,islam2024eob}. The idea of integrating XR consoles goes back to $1993$, with NASA's development of the first Telepresence ROV~\cite{hine1994application}, showcasing the feasibility of rendering camera feeds on an HMD. XR visuals have been utilized for various applications, including diver-ROV collaboration~\cite{domingues2012human}, VR-based jellyfish grasping~\cite{elor2021catching}, underwater cave exploration~\cite{abdullah2024ego2exo}, archaeological artifact reconstruction in the CoMAS project~\cite{bruno2018augmented}, and bathymetric mapping in the VENUS project~\cite{chapman2008virtual}. Candeloro~\etal~\cite{candeloro2015hmd} present a system where an HMD's head rotation sensing capabilities are utilized to control the ROV's roll, pitch, and yaw. 
%However, the use of the XR rendering device for direct ROV control remains underexplored. 
%However, their real-world tests focus on stabilizing the ROV for navigation and inspection, leaving its potential for more precise manipulation tasks largely unexamined. 

Besides vision, somatosensory (tactile and proprioceptive) feedback and control mechanisms are investigated to enhance the operator's sense of telepresence~\cite{dennerlein2000vibrotactile,lin2020compliant,xia2023visual,UnderwaterTactileSensorXu2022}. Recent studies demonstrate the utility of haptic cues delivered to the wrist, palm, and forearm region in providing insights into collisions~\cite{xia2024human,ryden2013advanced}, drag forces~\cite{stewart2016interactive}, and object stiffness~\cite{khedr2024design,lecuyer2009simulating} and their effects on operator performance. Additionally, kinesthetic feedback mechanisms have been developed to convey information about hydrodynamic states~\cite{xia2023sensory} including water currents~\cite{xia2023rov} and torsion forces~\cite{amemiya2009directional}. Research prototypes such as SeeGrip~\cite{kampmann2015towards} and modified versions of HaptX gloves~\cite{xia2023sensory} offer promising solutions but remain limited in portability for large-scale use. Research has also been conducted into developing fine underwater pressure sensors \cite{UnderwaterTactileSensorXu2022}. However, most studies on XR and haptic technologies are developed and evaluated in simulation environments~\cite{xu2021vr,zhou2023embodied}, leaving a critical gap in real-world testing and validation with subsea ROVs.
\section{System Design}

\subsection{Virtual Reality Interfacing}
The SubSense framework, equipped with a Newton Subsea gripper, is developed around the BlueROV2 Heavy system. In addition to its standard QGroundControl interface, we integrate a VR interface using the Oculus Quest 2. The system includes two hand controllers and an HMD, which streams real-time operational data from the BlueROV2, enhancing user immersion and situational awareness.

The BlueROV2 is equipped with a low-light HD camera that compresses live video into H.264 format for efficient transmission. The compressed stream is relayed via an onboard Raspberry Pi-4 to the host computer through a tethered network link. This video feed is then processed and displayed within the HMD using the Oculus Link, providing real-time visual feedback to the operator. We maneuver the BlueROV2 using an Oculus Quest-2 controller in the user's left hand. The grip trigger functions as a shift modifier, altering input functionality when fully depressed. In its default state, the joystick’s horizontal axis controls sway, while the vertical axis controls heave. The finger trigger commands forward surge. When the shift function is engaged, the horizontal joystick axis controls roll, the vertical axis adjusts pitch, and the finger trigger enables reverse motion.

Additionally, the HMD’s orientation is tracked to facilitate intuitive control. Yaw adjustments are made by tilting the head left or right, with thrusters actuating accordingly. The tilt angle determines the turn speed, up to a maximum thruster output. Similarly, HMD pitch is directly mapped to the servo-controlled camera mount, allowing the camera to rotate $\pm45^\circ$ for enhanced situational awareness.

    \begin{figure}[h]
        \centering
        \includegraphics[width=0.98\linewidth]{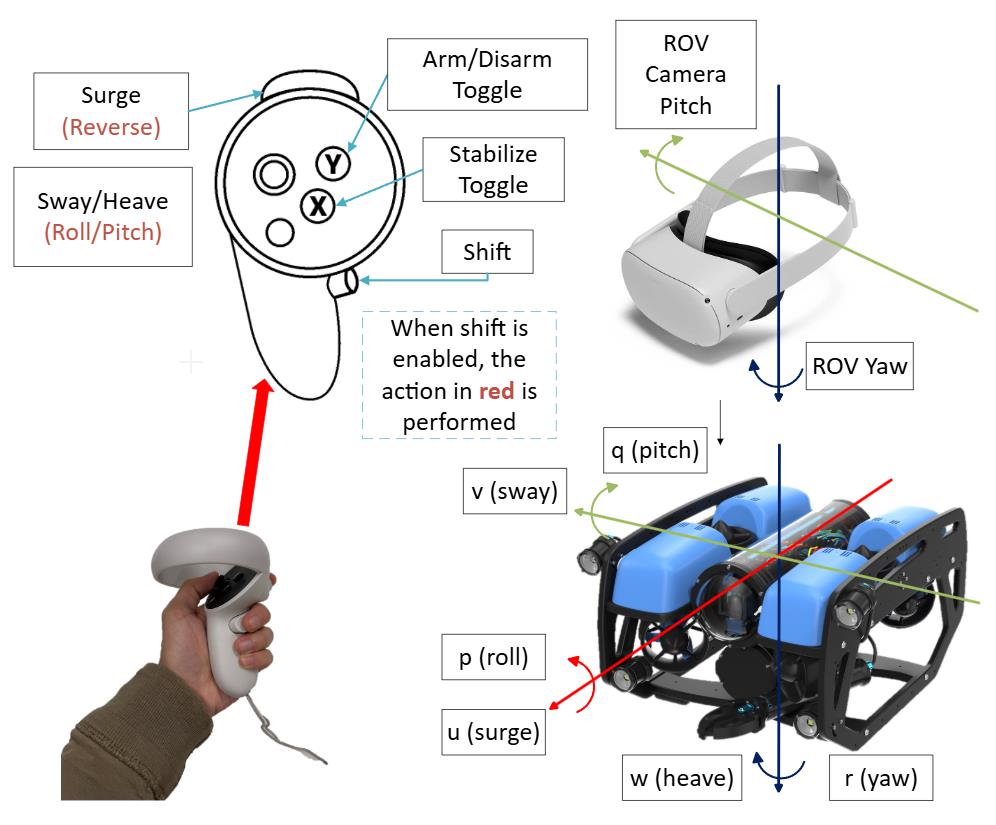}%
        \vspace{-1mm}
        \caption{The controller and HMD functions for ROV teletoperation by SubSense; the hand-held controller maneuvers the ROV while the HMD motions control \textit{yaw} and camera \textit{tilt}.}
        \label{fig:oculus_lctrl}
        \vspace{-3mm}
    \end{figure}
    
\subsection{Additional Software Integration}
A Docker container running a ROS2 Humble Ubuntu environment is used to translate user input into BlueROV2 actuation commands. Telemetry from the VR system is first acquired by the host computer and then transmitted to the ROS2 Docker. This data includes glove potentiometer positions, HMD position and orientation, and the position, orientation, and inputs from both controllers. The telemetry is published to separate ROS nodes, which process and translate the data into corresponding actuation commands. The resulting actuation commands are sent to the host computer, which transmits them to the BlueROV2 via the MAVLink protocol. The BlueROV2, in turn, returns telemetry data, processed and routed to the user in real-time to generate haptic feedback.

    % I could add a figure here detailing the yaw, the rate of change (from the IMU on board the bluerov), and seeing at what gain / coefficients that people feel the most comfortable at when turning the ROV.
    % I feel as though this is the point that may cause the most motion sickness.

\subsection{Haptic Interfacing \& Subsea Gripper} 
The integration of haptic feedback enhances user immersion during teleoperation. The BlueROV2 is equipped with a one-degree-of-freedom (1-DOF) manipulator, the Newton Subsea Gripper, which operates via pulse-width modulation (PWM) commands, similar to the ROV’s thruster control. The PWM commands are: $1500\mu s$ for neutral, $1530\mu s$-$1900\mu s$ for open, and $1470\mu s$-$1100\mu s$ for close. Due to the simplicity of this control scheme, direct glove-based manipulation is not feasible. However, position feedback is essential for mapping the gripper’s state to the operator’s hand movements, enabling more intuitive control.

\begin{figure}[h]
\vspace{-2mm}
    \centering
    \includegraphics[width=0.9\linewidth]{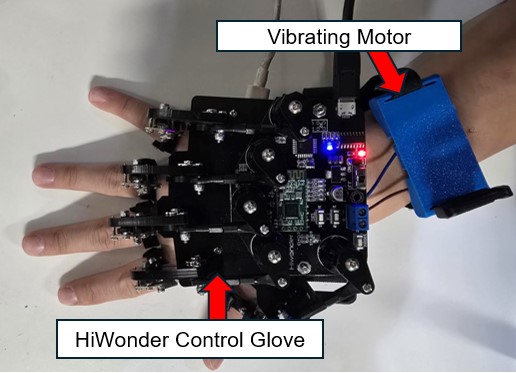}%
    \vspace{-1mm}
    \caption{The controlling glove for the right hand to actuate the ROV gripper with haptic feedback (vibration pattern) based on the force feedback of the ROV during grasping.}
    \label{fig:haptic_gloves} 
    \vspace{-1mm}
\end{figure}

\begin{figure}[t]
    \centering
    \includegraphics[width=0.98\linewidth]{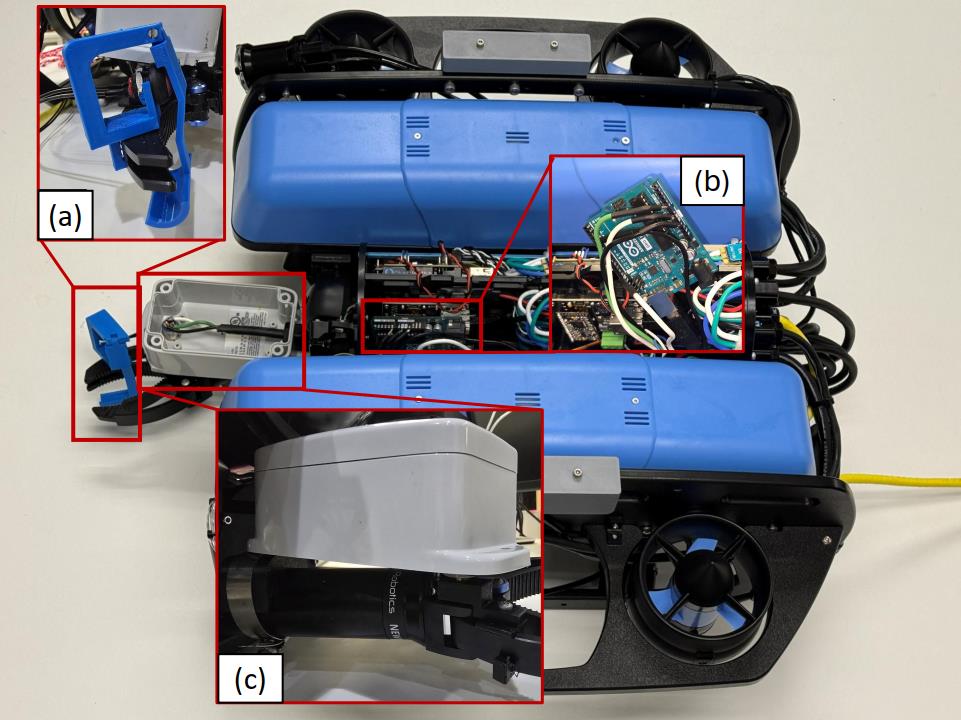}%
        \vspace{-1mm}
    \caption{Our non-invasive modifications to enable haptic feedback with: (a) a button with a paddle attachment; (b) an Arduino Uno; and (c) a potentiometer in a box.}
    \label{fig:ROV_Setup}
    \vspace{-3mm}
\end{figure}

To enable position control of the gripper, a linear potentiometer enclosed in a watertight casing is used to directly measure its rotation. While this provides positional feedback, it does not indicate when an object has been grasped. To address this limitation, a binary feedback mechanism is introduced by attaching a button to a 3D-printed mount on the gripper. To transmit potentiometer and button states to the host computer, an Arduino inside the ROV collects sensor data and communicates serially with the onboard Raspberry Pi. A serial-to-UDP bridge, implemented using BlueOS, transmits this data across the tether for real-time processing. Figure~\ref{fig:ROV_Setup} illustrates how these modifications integrate seamlessly with the existing BlueROV2 system.

We then use a HiWonder mechanical glove to directly map the user's hand position to the gripper pose. The glove is equipped with potentiometers that measure the contraction of each finger. By recording the open and closed positions of both the glove and gripper, intermediate positions are mapped linearly to enable intuitive control. The glove data is transmitted serially via a cable connection to the host computer, where both the glove and gripper data streams are processed to generate gripper actuation commands and haptic feedback signals. Haptic feedback is produced through a piezoelectric haptic motor, housed within a wrist-mounted attachment, as illustrated in Figure~\ref{fig:haptic_gloves}. An ESP32 microcontroller, receiving data from the host computer, generates the necessary PWM signal to actuate the motor. When the gripper’s feedback button is pressed, the motor produces a vibration for 2 seconds or until the button is released.

\subsection{Gripper Controller Design}
At the start of each use session, the glove calibrates to the user's natural \textit{open} and \textit{closed} hand positions, enabling personalized and comfortable control. Both glove and gripper positions are normalized on a scale from $0$ (fully open) to $1$ (fully closed), ensuring consistent and proportionate mapping. A `bang-bang controller'~\cite{Rusli2020_BangBang} is used to minimize the difference between glove and gripper positions.

%% Gripper issues not moving like expected %%%
% Mavlink commands
% Reference figure showing jumps in dy as well as discuss reasons for jumpiness

The gripper is controlled via ArduPilot using the MAVLink protocol. However, the API only supports discrete \textit{open} and \textit{close} commands, with no option to specify an intermediate position. This limitation results in large, inconsistent gripper movements, which vary depending on the gripper's current state. To enhance control precision, a neutral command is sent shortly after a move command to halt the gripper, with the timing between these commands directly influencing its behavior.

\begin{figure}[t]
    \centering
    \includegraphics[width=0.95\linewidth]{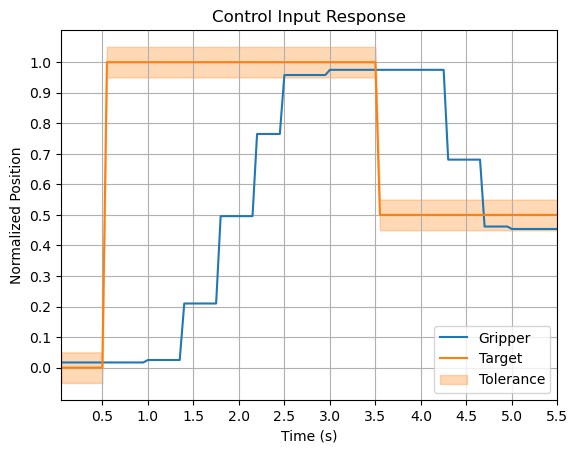}%
    \vspace{-1.5mm}
    \caption{Convergence of our gripper position control algorithm with input step responses. The orange line indicates the target (current hand position), while the blue line represents the gripper position. These results are from physical testing with the gripper using simulated hand positions.}
    \label{fig:ContResp}
    \vspace{-5mm}
\end{figure}

\begin{figure}[htbp]
    \centering
    \begin{subfigure}[]{0.48\textwidth}
        \includegraphics[width=\linewidth]{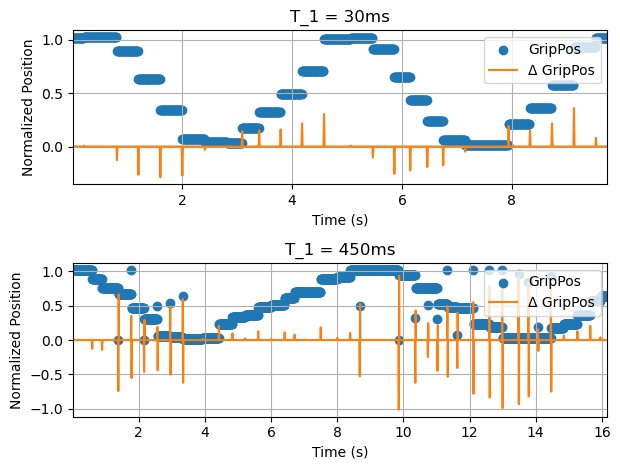}%
        \vspace{-1mm}
        \caption{Test for the time delay $T_1$ while $T_2$ is held constant at $45$\,ms.}
        \label{fig:SmallSleep}
    \end{subfigure}
    \vspace{1mm}
    
    \begin{subfigure}[]{0.48\textwidth}
        \includegraphics[width=\linewidth]{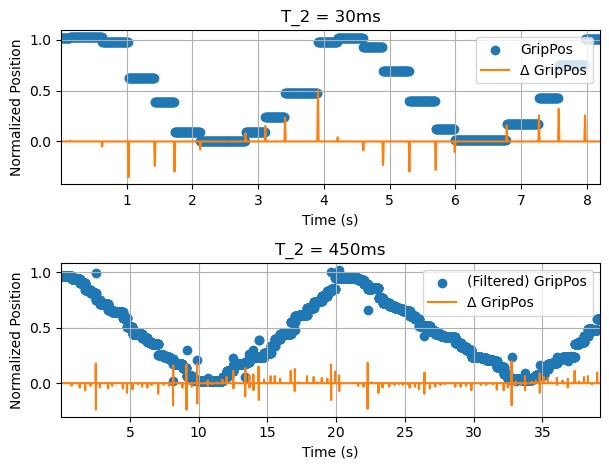}%
        \vspace{-1mm}
        \caption{Test for the time delay $T_2$ while $T_1$ is held constant at $45$\,ms.}
        \label{fig:BigSleep}
    \end{subfigure}
    \caption{Evaluations for the gripper operation and the change in movement with time for $T_1$ and $T_2$ independently.}
    \label{fig:TimeTest}
    \vspace{-3mm}
\end{figure}

We evaluate two timing variables, as shown in Figure~\ref{fig:TimeTest}: $T_1$, the delay between the move and neutral commands, and $T_2$, the delay between successive move commands. Testing reveals that shorter delays result in greater changes in grip position, while longer delays introduce abrupt movements, increasing noise in position measurements. Comparing their effects, increasing has a more pronounced impact on gripper speed than $T_1$. Consequently, subsequent tuning prioritizes $T_2$ to enhance control precision.

To achieve smooth and precise gripper control that accurately reflects user intent, the system is tuned to operate in an underdamped manner, minimizing overshooting (overgripping) and reducing the risk of object damage. Despite this tuning, some variability in gripper movement remains, as evidenced by the noise in Figure~\ref{fig:TimeTest}. To accommodate minor imprecision, we introduce a noise tolerance ($N_{tol}=0.05$ by default). We then map the delay $T_2$ using a function of the absolute normalized error $[0,1]$ which is the distance between the gripper and glove controller position, bounded between $[10,300]$\,ms. At larger error values, smaller $T_2$ values have larger and smoother movements, and vice versa. The control algorithm's step response behavior is illustrated in Figure~\ref{fig:ContResp}.

\section{Experimental Evaluation}
%The teleoperation framework is evaluated in a laboratory environment using a BlueROV2 outfitted with the augmented one-DOF gripper with haptic feedback arm to solve a three-disc Tower of Hanoi (ToH) problem. The evaluation focuses on metrics including completion time as well as the incidence of errors, external interventions, and catastrophic failures. Furthermore, operator feedback is collected to assess user comfort and preferences for control modes.

\subsection{Tower of Hanoi (TOH) Workspace}
The three-disc Tower of Hanoi (ToH) is a classic mathematical puzzle involving three poles within a given environment. Initially, a set of three stacked discs, arranged in increasing size from top to bottom, is placed on one pole. The objective is to transfer all three discs to a target pole while adhering to the following rules: (1) Only one disc may be moved at a time; (2) A disc can only be moved from the top; (3) A larger disc cannot be placed on a smaller disc; and (4) All discs not being moved must remain on a pole.

The ROV operates within a $10' \times 6'$\, water tank ($5'$ deep), where three 1-foot-tall towers are positioned approximately 2 feet apart. Each disc is composed of three 3D-printed components: a base, a cover, and a connector that snap together to form a complete unit. The base and cover feature centrally located circular holes, allowing the connector—designed with a wider base than the hole—to slide through, providing a secure gripping mechanism. The discs are arranged in increasing diameter from top to bottom, with the smallest disc at the top. Notably, the top disc has the largest center hole, while the holes in the larger discs progressively decrease in size. This design facilitates smooth stacking and ensures the discs easily slide into place when dropped during evaluation.

\subsection{Baseline Protocol and Evaluation}
The 3-disc Tower of Hanoi (ToH) problem consists of six "pick-and-place" sub-tasks. User performance is assessed from two perspectives: a first-person view (FPV) setup, serving as the baseline, and a VR + haptic system for comparative analysis. Each participant is allotted 30 minutes per perspective to complete the task. Performance is evaluated based on either the total completion time or the number of sub-tasks successfully completed within the allotted time. Additionally, we define the following supplementary metrics to facilitate a detailed comparison between the two perspectives:
\begin{itemize}%[label={$\bullet$},nolistsep,leftmargin=*]
\item \textbf{Minor damage}: damages that occur to the disk during manipulation, but do not impede task completion.
\item \textbf{Major Damage}: discs are damaged to the point that replacement of the part is necessary to continue.
\item \textbf{Collisions}: operator collides with objects or with the workspace boundaries. These are distinct from collisions involving the target object being manipulated, indicating a loss of spatial awareness of surroundings.
\item \textbf{Interventions}: instances where no damage occurs to a disc but is rendered inaccessible, requiring external help to place it in an accessible position. These involve resetting the discs or experimental setup altogether.
\end{itemize}

\begin{figure}[h]
\vspace{-1mm}
    \centering
    \includegraphics[width=0.96\linewidth]{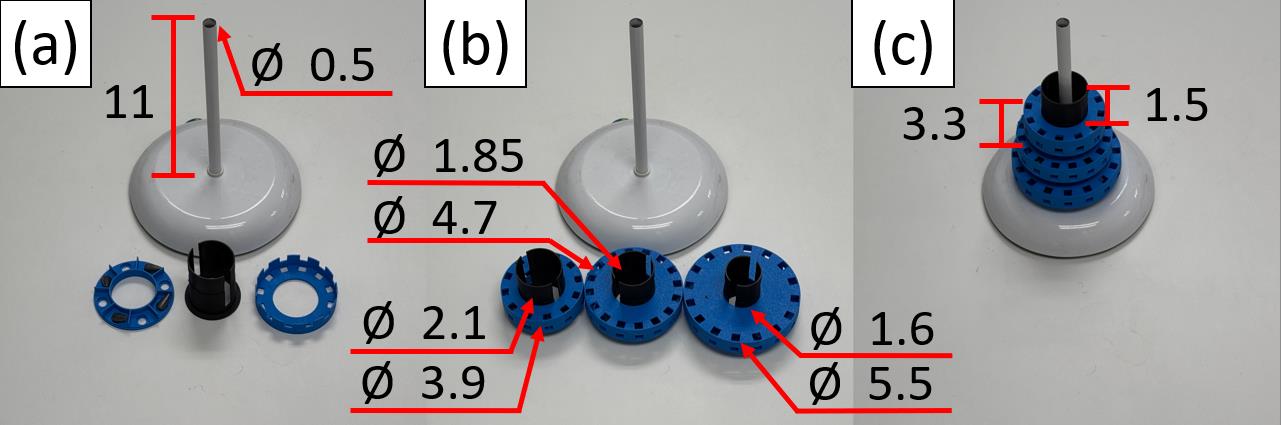}%
    \vspace{-1mm}
    \caption{The ToH discs used in our experiments: (a) a single disassembled disk, (b) three disks spread out, and (c) all three disks assembled on a tower; all dimensions in inches.
    }
    \label{fig:ToH_DiscSetup}
    \vspace{-2mm}
\end{figure}

%1. minor damage 2, major damage, 3, interventions, 4, wall/structure collisions
%For each perspective we total the amount of minor damages, which we define to be damage that has occurred to the disc but is otherwise still manipulable, and the amount of major failures which are when the disc is too damaged to the point where it is unusable, which also ends the evaluation. Interventions are defined as instances where no damage occurs but the disc is rendered inaccessible, requiring external help to replace it in an accessible position. To avoid damage to the ROV, the user is warned when a ROV thruster is about to collide with a tower. Collisions with towers and the walls of the workspace are also noted.

%\subsection{User Study}
%After completion of each experiment, each of our eight participants is asked to complete a user survey to gauge the usability and assess qualitative feedback for each system. While the system was found to be immersive, it can be seen that there were several attempts Table~\ref{tab:vr_perspective_comparison} that had to be stopped while the trial was underway due to disorientation from the VR headset. Noticeably, from observation of the users during both modes of operation, usage of VR elicits more careful and deliberate actions. This is corroborated by the differences in the amount of collisions from each participant in Tables~\ref{tab:fpv_perspective_comparison} and \ref{tab:vr_perspective_comparison}.

\begin{figure}[h]
    \centering
    \includegraphics[width=0.96\linewidth]{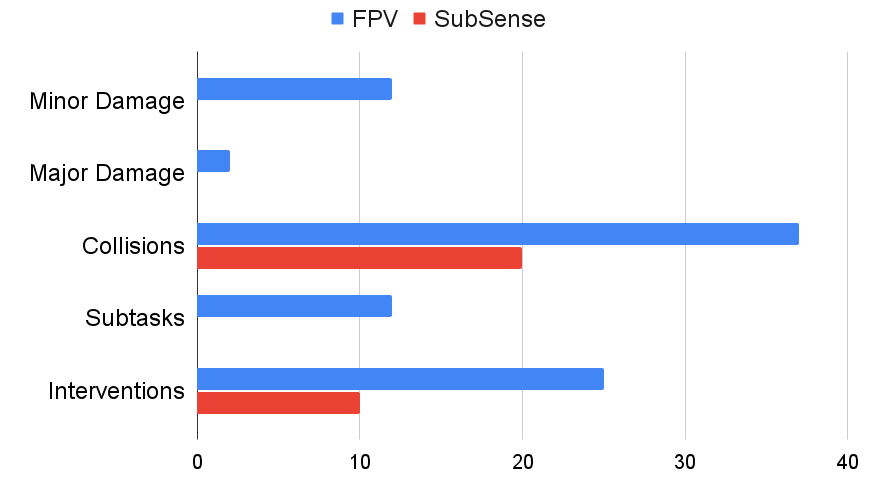}%
    \vspace{-2mm}
    \caption{Aggregated metrics from all trials using the traditional teleoperation control and through SubSense.}
    \label{fig:usergraph}
    \vspace{-4mm}
\end{figure}

After completion of each experiment, each of our eight participants is asked to complete a user survey to assess feedback for each system. Noticeably, from observation of the users during both modes of operation, usage of VR elicits more careful and deliberate actions. This is corroborated by Fig.~\ref{fig:usergraph}, which shows the significant drop in collisions or interventions required during the tasks. However, while SubSense was found to be immersive, several participants ended the test early due to disorientation.

\begin{figure*}[t]
\centering
	%\subfloat [The left shows an ROV with a gripper being controlled through SubSense using an alternative glove controller. On the right is an in-progress glove controller. Force feedback and finger position tracking are achieved through wires anchored to fingertips, connected to servos and potentiometers mounted on a forearm platform. A wrist-mounted PCB handles data processing and enables wireless communication and hand position tracking.] 
    \subfloat [An early version of the glove shows force feedback and finger tracking through fingertip-anchored wires linked to servos and potentiometers. A wrist-mounted PCB manages data processing, wireless communication, and hand pose tracking.] 
    {%
        \label{fig:future_hand}
		\includegraphics[ width=0.58\textwidth]{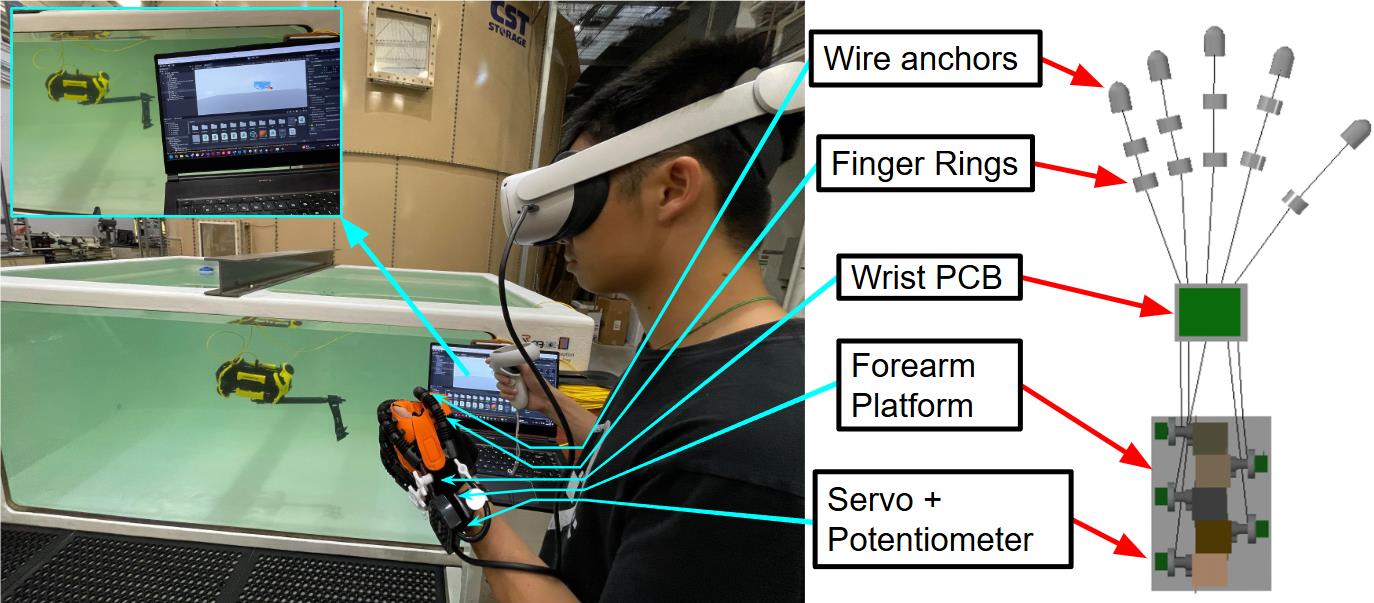}}~
        \subfloat[An interface mock-up featuring various overlays of detected objects, a minimap showing the positions of marked objects, and other relevant information.] {%
        \label{fig:interface}
		\includegraphics[width=0.38\textwidth]{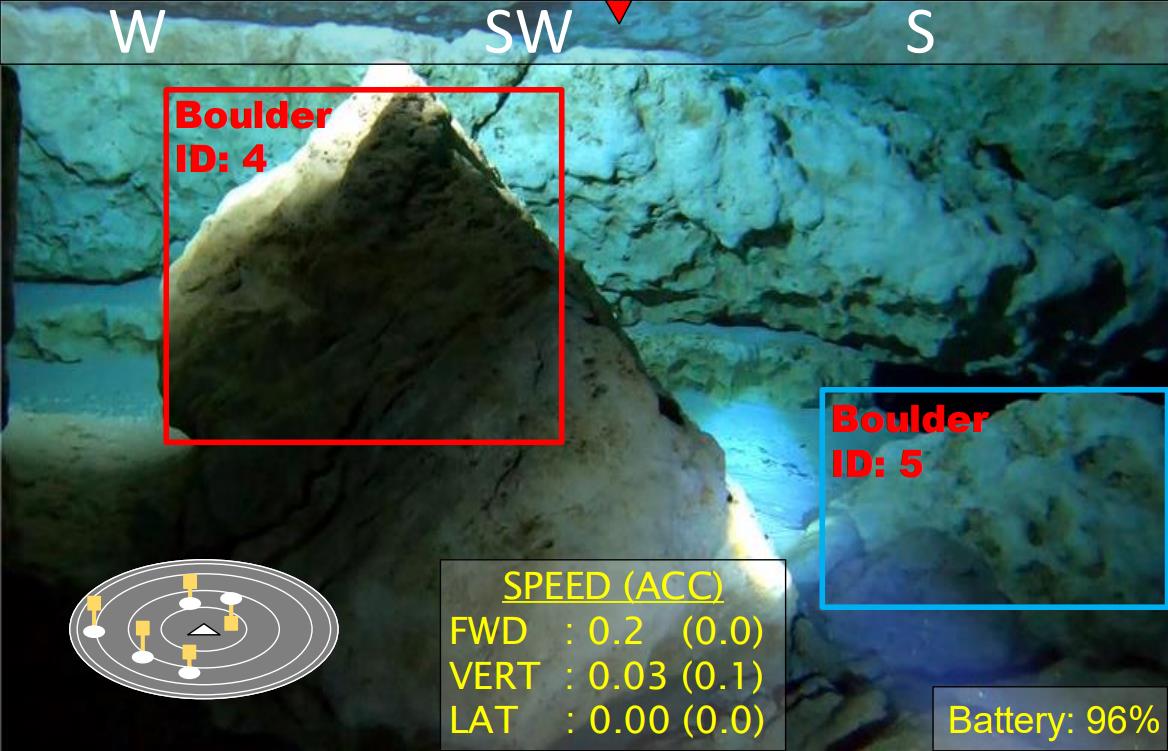}}
    %\vspace{2mm}
	\caption{Examples of ongoing works to upgrade our finger actuator glove and enhance the HMD UI with interactive features. %The left figure shows the glove controller currently being designed as well as exploring additional platforms for SubSense, and the right shows a possible HMD UI for a teleoperator while using SubSense. 
    }
 \label{fig:ongoing}
 \vspace{-3mm}
 \end{figure*}

\section{Limitations and Ongoing Work}
Initial findings from our user study highlighted several areas for improvement in the SubSense framework. Participants reported a noticeable increase in latency after approximately 10 minutes of continuous operation, leading to delays in both visual and control feedback, which negatively affected performance. While vibrational feedback was found to be beneficial, it was deemed insufficient for conveying detailed tactile information required for precise manipulation tasks.

Further limitations included the reliance on cabled connections for the haptic glove and HMD, which participants found restrictive and uncomfortable, limiting their range of motion. Additionally, disorientation occurred in featureless environments, making spatial awareness and navigation more challenging. While head-mounted yaw and camera tilt control were perceived as natural, joystick-based ROV movement control was described as unintuitive and disconnected, which hindered operational efficiency. To address these limitations, we are pursuing several enhancements to the framework.

\vspace{1mm}
\noindent
\textbf{F1.~Reducing latency}: One contributing factor to the observed latency is the communication overhead introduced by the Docker container. Presently, messages are routed through the container running the ROS nodes to generate commands for the ROV. Although this configuration enhances system portability, a more streamlined data flow could significantly reduce communication delays.

\vspace{1mm}
\noindent
\textbf{F2.~Glove controller upgrades}: The current glove controller provides only vibrational feedback and relies on a wired connection for data transmission to/from the host computer, limiting user mobility. We are developing an improved glove controller with integrated force sensing, shown in Fig.~\ref{fig:future_hand}, which enables users to experience physical feedback when interacting with objects. The updated glove will be battery-powered for wireless communication, enhancing flexibility. We are also refining the manipulator to incorporate position sensing, facilitating more responsive control.

\vspace{1mm}
\noindent
\textbf{F3.~ROV control improvements}: Advanced control algorithms can enhance the intuitiveness of ROV maneuvering by reducing reliance on joystick-based control. By integrating ROV position estimation and implementing a point navigation algorithm \cite{Ng2024_ArduSubMods}, the system can dynamically link the position and orientation of both the HMD and the ROV. This approach enables a more immersive user experience by eliminating the disconnect associated with using an external controller for vehicle movement, thereby enhancing operator embodiment and control precision.

\vspace{1mm}
\noindent
\textbf{F4.~Additional UI elements}: Certain sensory feedback elements, such as the resistance of water that naturally aids in perceiving speed and acceleration, are absent in the current system. To compensate for this limitation, it is essential to provide additional sensory cues to the operator. As shown in Fig.~\ref{fig:interface}, one approach is to overlay key metrics—including acceleration, velocity, and orientation—onto the camera feed within the HMD. Furthermore, incorporating additional visual elements, such as highlighting detected objects or proximity warnings, can enhance situational awareness and improve the operators' performance. We are in the process of developing and validating these features through field experimental trials in subsea inspection and mapping tasks.

%We plan to conduct more extensive user studies involving a larger number of participants over extended periods. In parallel, the user surveys will be refined to more effectively quantify user comfort and experience when operating the system. Additionally, we aim to evaluate the framework further in cave, lake, and oceanic settings. 

\section{Conclusion} \label{sec:conclusion}
This paper presents SubSense, a novel framework that integrates haptic feedback with virtual reality to enhance marine telerobotic operations. By combining these technologies, SubSense offers an immersive and intuitive control interface that improves operator situational awareness, efficiency, and safety in underwater environments. The framework enables seamless control of the BlueROV2 using commercial HMD devices, allowing users to visualize the live camera feed while intuitively maneuvering the vehicle. Additionally, we introduce a non-invasive modification to an off-the-shelf subsea gripper, providing real-time position and grasp status feedback. This enhancement allows for more precise manipulator control through a haptic glove system. A user study and performance evaluation demonstrate the potential benefits of SubSense over conventional interfaces, highlighting its promise for subsea telerobotics applications.

%While the system in its current iteration is lacking in robustness and long-duration usability, it provides valuable insights and a proof-of-concept foundation for more advanced, immersive teleoperation frameworks in future subsea applications.



\section*{ACKNOWLEDGMENT}
This work is supported in part by the University of Florida (UF) research grant \#$132763$ and the National Science Foundation (NSF) award \#$2330416$. 

%%%%%%%%%%%%%%%%%%%%%%%%%%%%%%%%%%%%%%%%%%%%%%%%%%%%%%%%%%%%%%%%%%%%%%%%%%%%%%%%

{\small
\bibliographystyle{IEEEtran}
\bibliography{survey_refs,other_refs,robopi_pubs}
}

\addtolength{\textheight}{0cm}   % This command serves to balance the column lengths on the last page of the document manually. 

\end{document}